\documentclass{article}
\usepackage[utf8]{inputenc}
\usepackage{amsmath}
\usepackage{amssymb}
\usepackage{amsfonts}
\usepackage{amsthm}
\usepackage{enumerate}
\usepackage{authblk}
\usepackage[colorlinks,linkcolor=blue,anchorcolor=blue,citecolor=blue,bookmarks = True]{hyperref}
\usepackage{longtable}
\usepackage{multirow}
\usepackage{float}
\usepackage{subfigure}
\usepackage{caption}
\usepackage{booktabs}
\usepackage{graphics}
\usepackage{graphicx}
\usepackage[margin=25mm]{geometry}

\usepackage{natbib}
\setcitestyle{authoryear,open={(},close={)}}
\usepackage{abstract}

\usepackage{graphicx}
\usepackage{verbatim}

\providecommand{\keywords}[1]
{
  \small    
  \textbf{\textit{Keywords:}} #1
}

\begin{document}
\title{A Dynamic Model for Traffic Flow Prediction Using Improved DRN}
\author[a]{Zeren Tan}
\author[b]{Ruimin Li}
\affil[a]{Department of Civil Engineering, Tsinghua University, Beijing 100084, China}
\affil[b]{Institute of Transportation Engineering, Tsinghua University, Beijing 100084, China}
%\thanks{lrmin@tsinghua.edu.cn}

\date{May 2018}

\maketitle
\begin{abstract}
    Real-time traffic flow prediction can not only provide travelers with reliable traffic information so that it can save people's time, but also assist the traffic management agency to manage traffic system. It can greatly improve the efficiency of the transportation system. Traditional traffic flow prediction approaches usually need a large amount of data but still give poor performances. With the development of deep learning, researchers begin to pay attention to artificial neural networks (ANNs) such as RNN and LSTM. However, these ANNs are very time-consuming. In our research, we improve the Deep Residual Network and build a dynamic model which previous researchers hardly use. We firstly integrate the input and output of the $i^{th}$ layer to the input of the $i+1^{th}$ layer and prove that each layer will fit a simpler function so that the error rate will be much smaller. Then, we use the concept of online learning in our model to update pre-trained model during prediction. Our result shows that our model has higher accuracy than some state-of-the-art models. In addition, our dynamic model can perform better in practical applications.
    
\end{abstract}
\section{Introduction}

Precise, quickly and timely traffic flow prediction is one of the major tasks of intelligent transportation systems (ITSs). It is of practical significance for individuals, companies and governments to make decisions according to real-time traffic flow. However, accurate and short-term traffic flow prediction remains challenging to researchers for decades because of its stochastic and nonlinear characteristics. At the beginning of traffic flow forecasting research, researchers mainly used linear methods, such as autoregressive integrated moving average (ARIMA) \citep{ARIMA}. Some researchers still prefer to use linear models for their simplicity and convenience and researchers have proposed some improved linear models such as multi-variable linear regression (MVLR) \citep{MVLR}. 
\par
From about 40 years ago, some machine learning algorithms showed good performance in many tasks, researchers began to use machine learning algorithms like support vector regression (SVR) \citep{SVR} and k-Nearest Neighbor (k-NN) \citep{k-NN}. In spite of good performance, these approaches cannot consider the entire characteristics in traffic flow and do not have satisfactory performances.
\par
Recently, with the development of deep learning, some deep learning models for traffic flow prediction are put forward, such as Recurrent Neural Network (RNN) \citep{RNN}, Long Short-Term Memory network (LSTM) \citep{LSTMtraffic}, Gated Recurrent Unit (GRU) \citep{gruTraffic}, Stacked Auto-Encoders (SAEs) \citep{SAE}, Deep Belief Network (DBN) \citep{DBN}, etc.. In general, these models have complex network architecture that can capture nonlinearities in traffic flow. Hence, they perform better in forecasting traffic flow than traditional models. Despite of this, models can be improved in many ways and performance can be even better.
\par
Among all these deep learning models, RNN \citep{RNN}, LSTM \citep{LSTMtraffic} and GRU \citep{gruTraffic} are the most commonly used models. However, the architecture of these models is so complicated that it takes a lot of times to train them. Since it is hard to train, people often have to stack less layers and set the training epochs to a small number, which will result in lower accuracy, or people may use some techniques such as dropout to reduce the size of training set and test set which does not allow people to get a higher accuracy when the data set is not large enough. Therefore, it is difficult to use these models in practical applications. On the other hand, previous studies were mainly based on static models, that is, the model will not be updated once it was trained. Nevertheless, traffic flow is the trick of the real world. In addition, the spatial and temporal relationship of traffic flow is is changing all the time. Therefore, we need to build the mechanism of real-time updating. 
%\par
%\citet{DRN} proposed deep residual network (DRN) in 2016.
%DRN has showed excellent performance in computer vision competitions. What is important is that DRN cost much less time to be trained than RNN, LSTM and GRU without the expense of higher error rate. However, researchers haven’t applied DRN in traffic flow prediction. We are inspired by the ideas behind DRN and attempt to apply it in traffic flow prediction. In order to apply it in our field, we improve the architecture of DRN and make it more powerful.
 \par
In this paper, we improved the deep residual network (DRN) and proposed a Dynamic Improved Deep Residual Network (DIDRN) that will continuously update its training set when some real data are available, which is more powerful in practical application. The DIDRN has good adaptability when the road conditions and the places where we apply it change over time.

%To summarize, our main contributions are as follows:
%\begin{enumerate}[(1)]
%    \item We introduce DRN and improve it to predict traffic flow.
%    \item We propose a dynamic deep neural network called DIDRN that is more powerful in practical application.
%    \item Our model is much easier to train and have a higher accuracy than other models such as RNN and LSTM.
%\end{enumerate}

The rest of this paper is organized as follows. Section  \ref{sec:LR} reviews the existing works on short-term traffic flow prediction. Section \ref{sec:DRN} demonstrates basic ideas of DRN. Section \ref{sec:improv} and Section \ref{sec:dyna} explain how we further improve DRN and how we design the dynamic model, respectively. Section \ref{sec:exre} develops experiments and compares our model's performance with several popular models. Finally, section \ref{sec:confw} concludes this paper and discusses future works.

\section{Literature Review}
\label{sec:LR}
Since traffic flow prediction is one of the main tasks of ITSs, researches on traffic flow prediction have been ongoing for many years. New models are constantly being proposed. 
Although there are numerous traffic flow prediction methods, generally, they can be divided into two main categories: parametric methods and non-parametric methods.
\par
Parametric methods such as ARIMA \citep{ARIMA} and MVLR  \citep{MVLR} are often used by researchers. These approaches require predetermined model architecture and that the parameters of the model are calculated by empirical data \citep{deeptrend}. Some improved models based on ARIMA like SARIMA \citep{SARIMA} were proposed in the 20th century. These models have simple and explicit architecture but require a huge amount of data and that the traffic condition is in a stationary process. Hence, it may be impossible to use these approaches when sufficient data are unavailable. 
In most cases, traffic flow data have stochastic and nonlinear features. Therefore, those parametric approaches cannot perform very well. 
\par
From the beginning of 20th century, researchers began to pay more attention to non-parametric approaches such as k-NN \citep{k-NN}, SVR \citep{SVR}, random forests regression (RF) \citep{RF} gradient boosting regression \citep{GB} and so on. On the other hand, due to the excellent performances of Artificial Neural Networks (ANNs) in capturing nonlinearities, researchers have paid more and more attention to them since the birth of ANNs. 
\par
Recently, with the rapid development of deep learning, a lot of deep learning models have been proposed by computer scientists. Many powerful deep learning models, such as SAE \citep{SAE}, DBN \citep{DBN}, RNN \citep{RNN}, LSTM \citep{LSTMtraffic} and GRU \citep{gruTraffic}, were introduced to traffic flow prediction and showed superior performances. 
\par
Sequential data are correlated with data in the context. Nevertheless, common neural networks cannot capture this correlation among data. In the 1980s, RNN \citep{RNNraw} was developed to solve this problem. However, although RNN can capture correlation among data, it will soon forget this correlation, that is, it does not have a long-term memory. 
In 1997, long short-term memory (LSTM) networks \citep{LSTM} were invented and set accuracy records in various applications domains. LSTM can remember correlation for a long time and thus has an outstanding performance in sequential modeling.
However, both RNN and LSTM are very hard to train since the architecture of these models are too complex. It often takes researchers hours or even days to train these models once. Although they have good performance, to use them in practice is sometimes impossible. 
\par
In order to reduce mathematical operations and total running time, gated recurrent unit (GRU) was developed by \citet{GRU}. Unfortunately, in practice, although GRU can reduce training time, the cost is still unacceptable. In addition, \citet{lstmcomparison} made a good comparison of the variants of LSTM and found that they are all the same.
\par
Besides RNN and LSTM, researchers have also tried some other deep learning models.
\par
\citet{ELM} proposed a novel double deep ELMs ensemble system (DD-ELMs-ES) and their model demonstrates better generalization performance than some state-of-art algorithms. However, their model only focuses on one-step forecasting.\citet{deeptrend} implemented DeepTrend which decomposes original traffic flow data into trend and residual components. They demonstrate that DeepTrend can noticeably improve the prediction performance and outperform many traditional models and LSTM. But they didn’t take the spatial correlation into account. \citet{DBEN} proposed a deep belief echo-state network (DBEN) to address the problem of slow convergence and local optimum in time series prediction. Experiments results demonstrate that DBEN has a good performance in learning speed and short-term memory capacity. However, there are a large number of parameters in their model so parameter tuning would be a really hard work.
\par
\citet{DRN} proposed deep residual network (DRN) in ILSVRC. Their model substantially outperformed other models and won the first prize in this competition. Typically, when networks become deeper, it will be difficult to train them and the problem of gradient vanishing will arise.  Their model is easy to train and can solve the problem of gradient vanishing effectively even if the deep neural network is several times deeper than other networks.  For a normal deep neural network, the maximum depth that people have ever used is less than 100. But for DRN, we can easily create a network with several hundred layers and train it in a even shorter time.

\section{Model Development} \label{sec:method}
\subsection{DRN}\label{sec:DRN}

\begin{figure}
    \centering
\begin{minipage}[t]{0.48\textwidth}
\centering
    \includegraphics[width = 6cm]{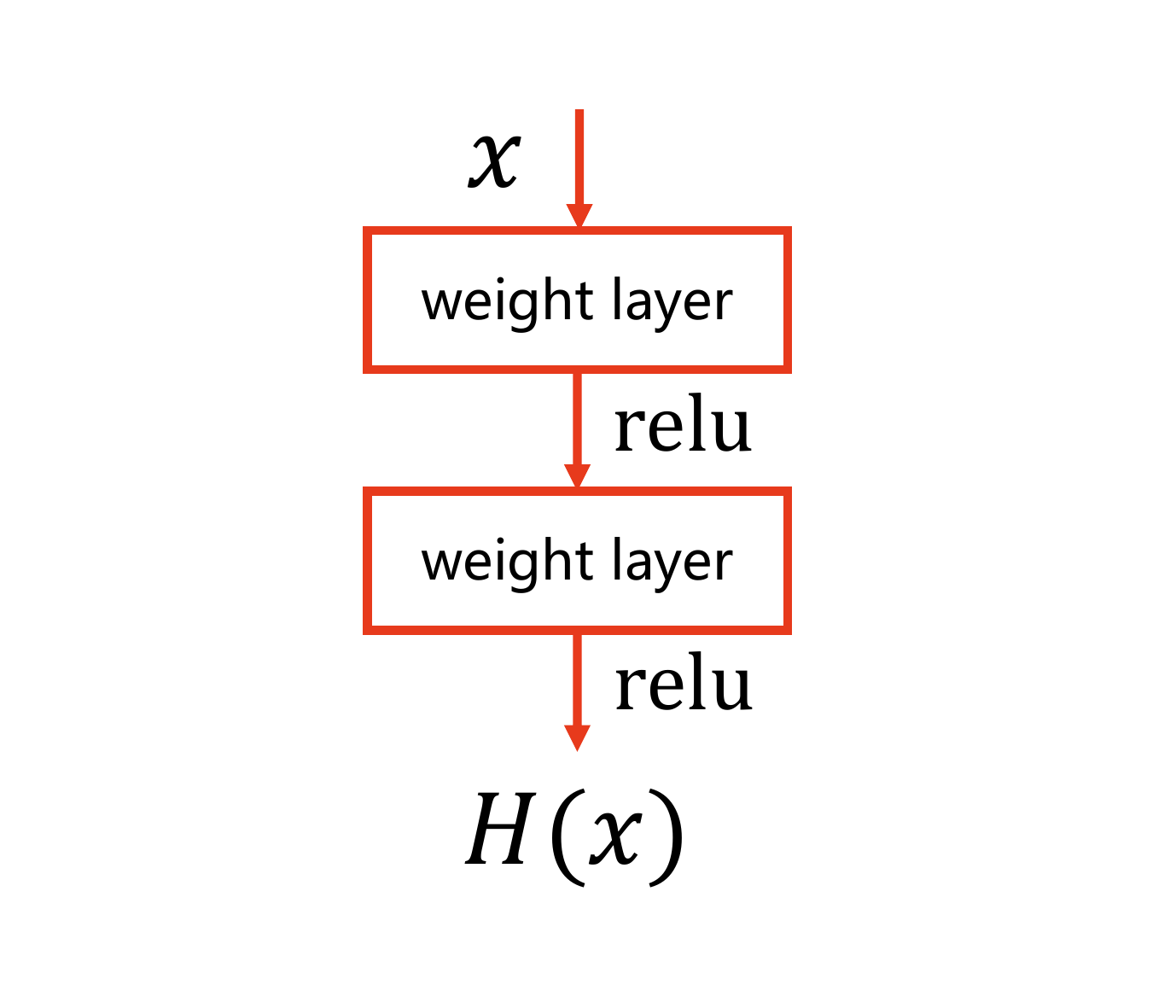}
    \caption{Architecture of common neural network}
    \label{fig:com_nn}
\end{minipage}

\begin{minipage}[t]{0.48\textwidth}
\centering
    \includegraphics[width = 6cm]{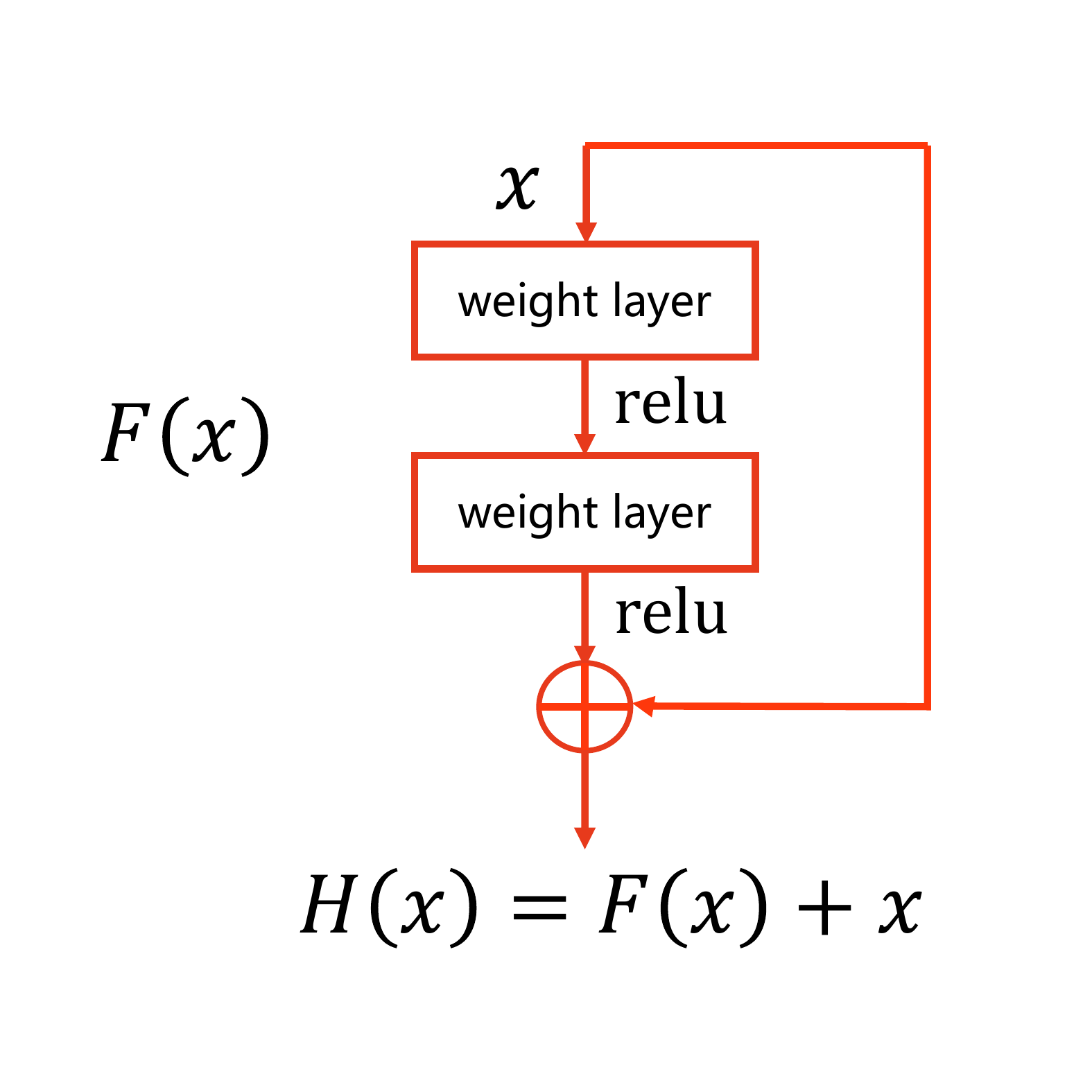}
    \caption{Architecture of residual neural network}
    \label{fig:res_nn}
\end{minipage}

\end{figure}

When deep learning began to boost, people simply stacked layers and expected that deeper is better. They give the network an input $x$ and let the intermediate layers fit a map $H$ to get the final output $H(x)$. As they believe deeper networks could have better performances, they stack more layers to fit the desired map $H$ and expect to get a higher accuracy. However, it turns out that when the depth of the network increases, both training and test accuracy get saturated, that is, degradation arises.\citep{DRN} Figure \ref{fig:com_nn} shows the architecture of simply stacked neural network.
\par
DRN is proposed to solve the problem of degradation. Instead of hoping each stacked layer fits a desired mapping directly. Kaiming lets some layers fit a residual mapping \citep{DRN}.
\par
Concretely, assuming that we want our network to fit $H(x)$, we let the sacked layer fit another mapping $F(x) = H(x) –x$ . Then the desired mapping $H(x)$ can be replaced by $F(x)+x$.
\par
We hypothesize that it is easier to fit the residual mapping $F(x)$ than the original mapping $H(x)$. If the optimal mapping is identity, then it would be easy to make the residual be zero. Figure \ref{fig:res_nn} shows the idea stated above.
\par
This simple change makes a lot of difference. The network becomes much easier to train. Deep residual network has helped Kaiming and his team win 1st prize in the ILSVRC classification competition and many other competitions. Before DRN existed, the winner network of ILSVRC had at most 32 layers while when DRN participated in this competition, it had 152 layers. What's more, the accuracy had a significant increase.

\subsection{Improved DRN} \label{sec:improv}
\begin{figure}
    \centering
    \includegraphics[width = 6cm]{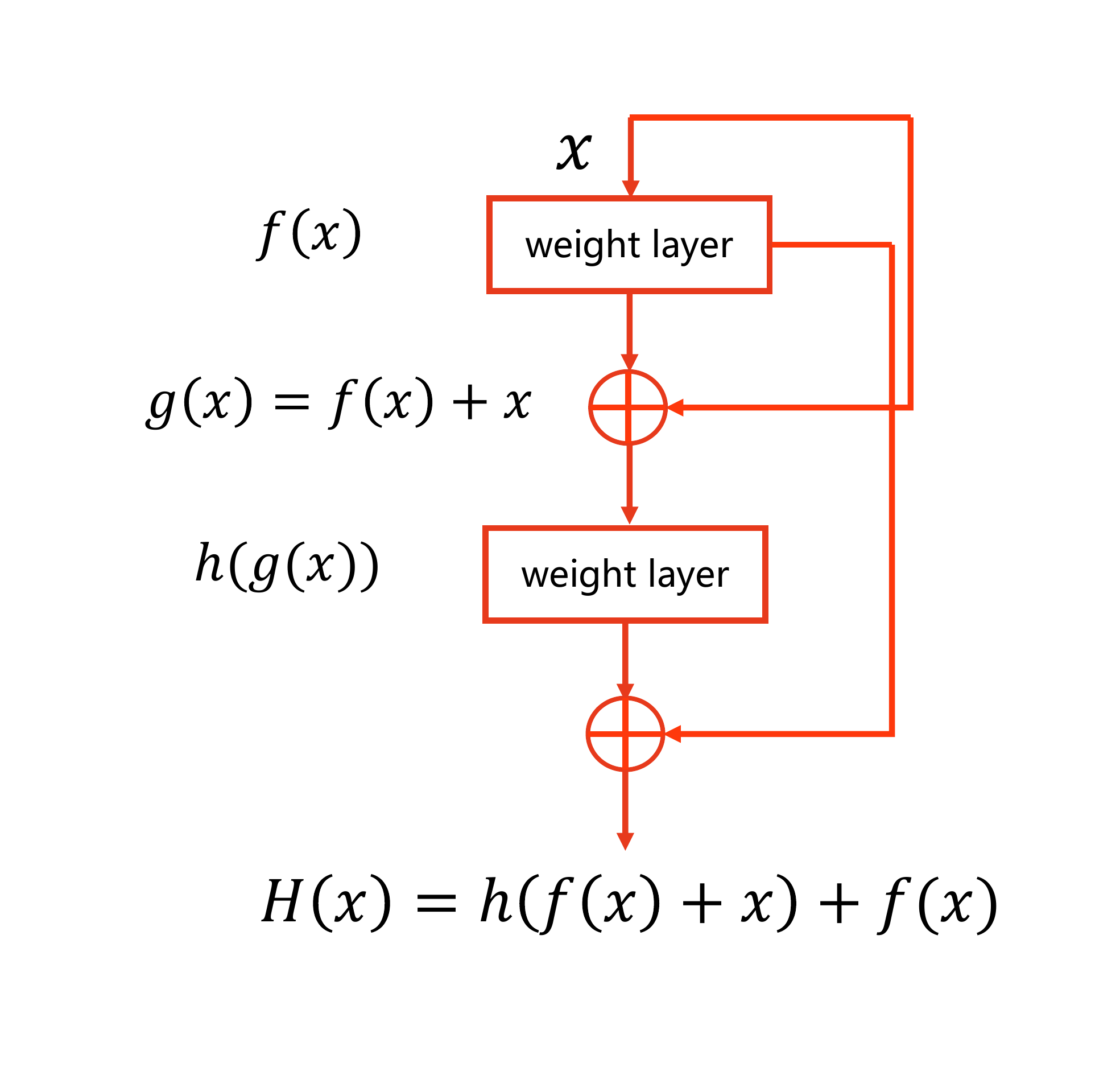}
    \caption{Our improved residual network}
    \label{fig:imp_nn}
\end{figure}

We are inspired by the idea of Kaiming on deep residual network. To keep more information when information is transfered among layers and thus get higher accuracy, we further improve the architecture of DRN and use it in traffic flow prediction. Figure \ref{fig:imp_nn} shows architecture of our improved network.

\par
The network works as follows: 
\begin{itemize}
\item Denote the input as $x$, and the first layer fits the function $f(x)$. 
\item Then we add these two outputs together and we get $g(x) = f(x) +x $. 
\item The second layer takes $g(x)$ as input and maps it into $h(g(x))$. 
\item Similarly, we then add the output of the first layer $f(x)$ and the output of the second layer $h(g(x))$ and we get $H(x) = h(f(x)+x) + f(x)$. 
\end{itemize}
\par 
From the aforementioned steps, it is clear that we divide the task of fitting a complex function $H(x)$ into fitting two simpler mappings $f$ and $h$, respectively. It will make the layers work more efficiently since it is easier to fit these two simple mappings. Consequently, it can reduce the error rate of fitting these two mappings and eventually reduce the overall error rate.

\subsection{Dynamic Model} \label{sec:dyna}
In practice, transportation management departments will not use this model in the specific place that our data come from, that is our model needs to be transfered easily among different places. Besides this, traffic flow on a specific road may change over time. In other words, the traffic flow after one month or one year may have significant difference with the current one. The weights and parameters were obtained by using the old training data. These weights and parameters can fit the old data very well but it can not fit the new data well, that is the previous model would not have good performances on new data. Therefore, it is unreasonable to simply use the old model to predict traffic flow without updating it. It is also not surprising that we cannot expect this old model to have good performances when the traffic flow has changed significantly. Hence, traffic flow prediction models should be updated constantly. Based on the ideas stated above, we choose to design a dynamic model.
\par
In order to update pre-trained model, we use the basic idea of incremental learning to implement a dynamic model. 
\par
The complete process is as follows:
\begin{enumerate}[Step 1]
\item\label{step:1} We use the collected data to pre-train a basic model.
\item\label{step:2} Then we use our model to do prediction work.
\item\label{step:3} When finishing some steps of prediction, we get some new real data.
\item\label{step:4} We combine these new data into a new training set and use this new training set to train our model again.
\item\label{step:5} Repeat Step \ref{step:2}-Step \ref{step:4}.
\end{enumerate}
\par
\begin{figure}
    \centering
    \includegraphics[width=8cm]{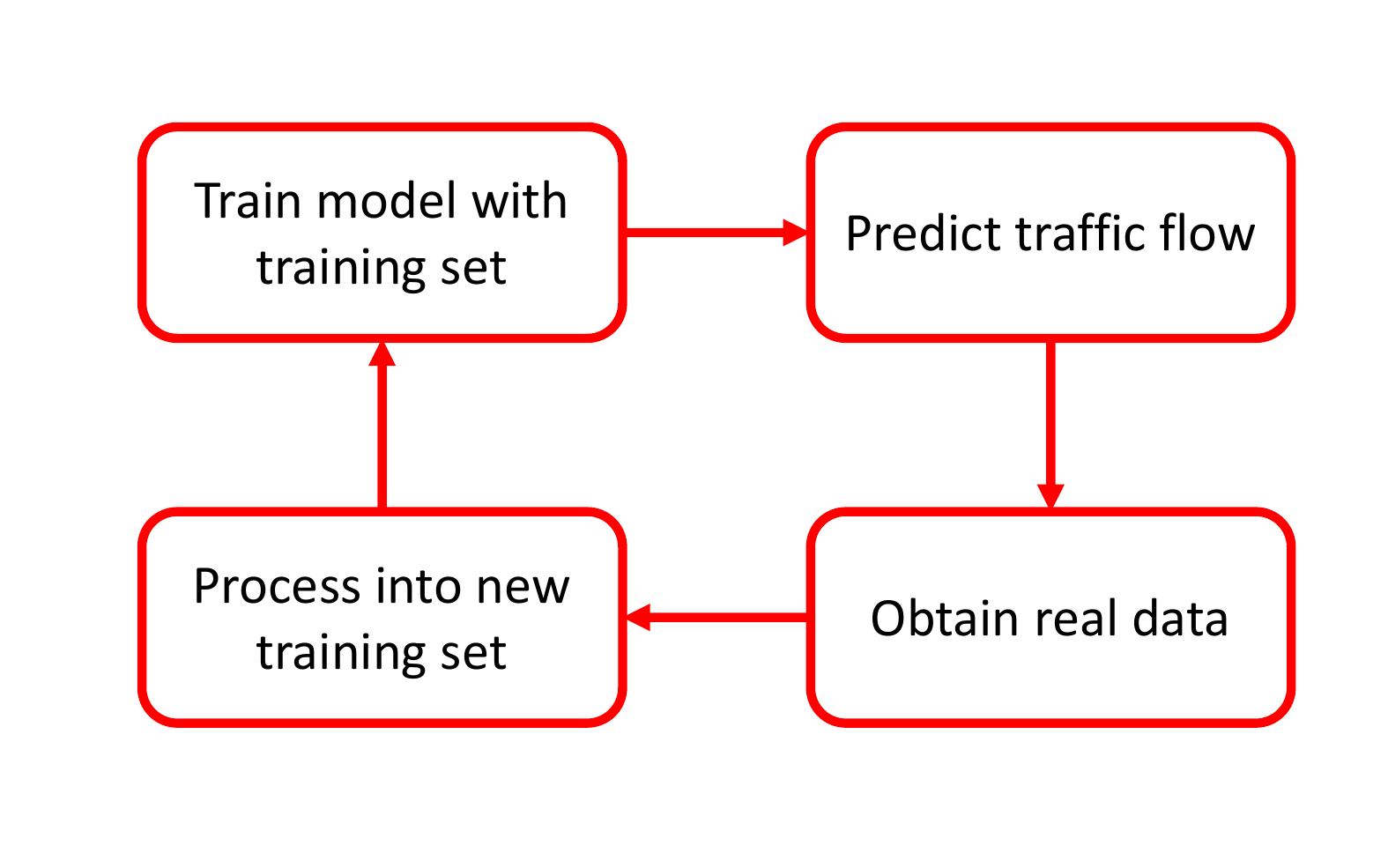}
    \caption{Dynamic Model}
    \label{fig:steps}
\end{figure}

Figure \ref{fig:steps} shows the process of these steps.

\par
With these steps, the model will be continuously updated through absorbing new data. Therefore, this model will fit the practical conditions better and give more accurate predictions.
\subsection{Model Implementation}
In this section, we will talk about the whole process of processing data.
\par
We divide the process into several steps:
\begin{enumerate}[Step 1]
\item\label{step:11} Denote the raw data as $R$. It is a column vector of $R^n$. 
\[
R=\begin{bmatrix}
R_1\\
R_2\\
\vdots\\
R_n
\end{bmatrix}
\]
\par
We reasonably assume that there are some trend characteristics in traffic flow data. So we subtract adjacent data to get the difference vector $D=[D_1, D_2, \cdots, D_{n-1}]$. Where
\[
D_i = R_{i+1}-R_i \quad i = 1,2,\cdots,n-1
\]
\item\label{step:22} However, we cannot apply supervised learning to $D$ directly. Hence, we further process our data into supervised data. 
\par
First, we ought to choose the time-step. Time-step represents how many previous data points we use to predict the next data point. After parameter tuning, we choose 1 as our time-step, that is, time-step = 1. 
\par
Then we denote $X$ as the feature and $Y$ as the data to be predicted. $X$ and $Y$ are defined as follows.
\[
X = \begin{bmatrix}
X_1\\
X_2\\
\vdots\\
X_n
\end{bmatrix}
\quad
Y = \begin{bmatrix}
Y_1\\
Y_2\\
\vdots\\
Y_n
\end{bmatrix}
\]
Where
\[
X_1 = 0\quad X_i = D_{i-1},i= 2,3,\cdots,n-1
\]
\[
Y_i = D_{i}, i = 1,2,\cdots,n-1
\]
\par
Then we get the entire data set.
\[
\begin{bmatrix}
X_1&Y_1\\
X_2&Y_2\\
\vdots & \vdots\\
X_{n-1} &Y_{n-1}
\end{bmatrix}
\]

\item\label{step:33} In order to accelerate the speed of training and prediction, a simple approach that people often use when developing deep learning model and doing experiments is normalization. Since some elements in $D$ are positive and others are negative, we scaled $X$ and $Y$ into $(-1,1)$. We define the resulted vectors as $scaled\_X$ and $scaled\_Y$. They are defined as follows:
\[
scaled\_X = \begin{bmatrix}
scaled\_X_1\\
scaled\_X_2\\
\vdots\\
scaled\_X_{n-1}
\end{bmatrix}
scaled\_Y = \begin{bmatrix}
scaled\_Y_1\\
scaled\_Y_2\\
\vdots\\
scaled\_Y_{n-1}
\end{bmatrix}
\]
\par
Where
\[
scaled\_X_i = 2\frac{X_i - X_{min}}{X_{max}-X_{min}}-1
\]
\[
scaled\_Y_i = 2\frac{Y_i - Y_{min}}{Y_{max}-Y_{min}}-1
\]
\par
And
\[ 
scaled\_X_i \in (-1,1), i = 1,2,\cdots,n-1
\]
\[
scaled\_Y_i \in (-1,1), i = 1,2,\cdots,n-1
\]
\par
Then the data set becomes
\[
\begin{bmatrix}
scaled\_X_1 & scaled\_Y_1\\
scaled\_X_2 & scaled\_Y_2\\
\vdots&\vdots\\
scaled\_X_{n-1} & scaled\_Y_{n-1}
\end{bmatrix}
\]
\item\label{step:44} When predicting, we invert Step \ref{step:11} and Step \ref{step:33} to get the final prediction data.
\end{enumerate}
\section{Experiments and Results} \label{sec:exre}
\subsection{Data Source}
Our data come from the ring roads in Beijing, China. Basic traffic flow data are collected by 14 microwave detectors. The interval of time of these traffic flow data is 10 minutes. 
The total length of data collected by each detector is 8784. The time span is 61 days from $1^{st}$ Jun., 2013 to $31^{st}$ Jul.,2013 . For each series of data, we use the first 7200 pieces of the entire data as training data and the rest as test data. We use the detected traffic flow as input and finally get the predicted traffic flow at specified time points. The results verify the feasibility and effectiveness of our proposed model.

\subsection{Performance Indexes}
In order to evaluate the performance of our proposed model, we develop several experiments on our model and several similar models, including one layer LSTM, deep LSTM, and DRN. We use one layer LSTM as the baseline model to show the accuracy promotion of our model. We use LSTM to illustrate that DIDRN outperforms some state-of-the-art and widely used models. Finally, DRN is included to demonstrate that the improvement we have made does make sense.
\par 
We use Rooted Mean Square Error (RMSE), Mean Absolute Percentage Error (MAPE) and Mean Absolute Error (MAE) to evaluate the performance of these four models.
\par 
RMSE, MAPE and MAE are defined as follows:
\[
RMSE = \sqrt{\frac{1}{n}\sum_{i=1}^{n}(\hat{y}-y)^2}
\]
\[
MAPE = \frac{1}{n}\sum_{i = 1}^{n}\frac{|\hat{y_i} - y_i|}{y_i} \times 100\%
\]
\[
MAE = \frac{1}{n}\sum_{i=1}^n|\hat{y}-y|
\]
\par
Where $y_i$ is the real value and $\hat{y_i}$ is the forecast value.
\par
RMSE and MAE represent the deviation from the predicted values and the detected values. They will expand as the range of the data is expanded. Therefore, we can not evaluate our model just on the basis of the absolute value of RMSE and MAE. The range of our data should be considered at the same time. However, MAPE is a relative measure of the deviation from the predicted values and the ground truth values. It is a percentage error so it is not related to the range of our data. We use these three performance indexes to evaluate two aspects of the selected models: 1) absolute error 2) relative error.

\subsection{Performance Analysis}
Except for one layer LSTM, the other three models have 16 layers. In order to change dimension, we add some layers in both DRN and DIDRN but they still have 16 layers in total.
\par 
We evaluate models by 14 different sets of traffic flow data from 14 detectors. They are numbered as 2010, 2011, 2013, 2023, 2030, 2033, 2052, 3034, 3035, 4004, 4005, 4050, 4051, 5062.
\par
All the experiments are run on a desktop with @2.60 GHz processor, 8.0 GB RAM and NVIDIA GeForce GTX 960M. All models were coded in Python 3.6 with Keras and Tensorflow framework and they were compiled by Anaconda Jupyter Notebook.
\par 
We first use our these model to predict short-term traffic flow. Concretely, the interval of time points is 10 minutes. We use the previous traffic flow to predict the traffic flow 10 minutes later.
\par
Table \ref{tab:1} shows the performance.

\begin{longtable}{cccccccccccccccc}
  %  \centering
    \caption{Performance comparison of different models}
    \label{tab:1}\\
   % \begin{tabular}{cccccccccccccccc}
    \toprule
        \multicolumn{2}{c}{Model} & One Layer LSTM & Deep LSTM & DRN & DIDRN \\
        \midrule
        \multirow{3}{*}{2010} &RMSE &  101.07 & 75.65& 75.64 & \textbf{74.35}\\
        \endfirsthead
        & MAPE &  11.58\%  & 7.05\%
  & 7.04\% & \textbf{6.99\%} \\
        & MAE & 84.60 & 59.13 & 59.12 & \textbf{53.73}\\
        \hline
        \multirow{3}{*}{2011} & RMSE & 123.79 & 104.13 &104.13 & \textbf{101.53}\\
        & MAPE  &11.15\% &7.90\% & 7.90\%
 & \textbf{7.44\%}
\\
        & MAE  & 98.61 & 77.70& 77.71 & \textbf{75.07}
\\
        \hline
        \multirow{3}{*}{2012} &RMSE  &122.05 &96.47 & 96.37 & \textbf{94.46} \\
        & MAPE  & 15.56\% & 9.46\%
& 9.43\% & \textbf{9.38\%}
\\
        & MAE  & 98.82 & 71.23& 71.13 & \textbf{69.07}
\\
\hline
        \multirow{3}{*}{2023} &RMSE& 99.99 &78.28 & 78.02 & \textbf{74.26} \\
        & MAPE &11.19\% & 7.02\%
& 6.97\% & \textbf{6.96\%}
\\
        & MAE & 81.86 &58.90 & 58.60 & \textbf{55.72}\\
        \hline
        \multirow{3}{*}{2030}&RMSE  &104.95 &76.00 & 75.96 & \textbf{67.13}\\
        & MAPE &14.63\%&7.43\% 
 & 7.42\% & \textbf{7.38\%}
\\
        & MAE & 86.94 & 57.34&  57.29 & \textbf{49.66}\\
        \hline
        \multirow{3}{*}{2033}&RMSE &79.84 &59.75 & 59.75 & \textbf{58.40} \\
        & MAPE & 12.63\% &6.89\%
 & 6.89\% & \textbf{6.80\%}
\\
        & MAE & 66.70 & 45.11& 45.11 & \textbf{44.11}\\
        \hline
        \multirow{3}{*}{2052} & RMSE & 109.56 & 85.99 
&85.99 & \textbf{83.28} \\
  & MAPE & 11.63\%& 7.46\%
&7.46\%  & \textbf{7.20\%} \\
& MAE & 88.31 & 65.52 & 65.52 & \textbf{62.77} \\
 \hline
\multirow{3}{*}{3034} & RMSE & 99.08 
 & 74.32 & 74.04 & \textbf{72.27} \\
& MAPE & 10.45\% &6.97\%  &6.89\% & \textbf{6.89\%} \\
& MAE &83.05  & 57.07 & 56.99 & \textbf{53.67} \\
\hline 
\multirow{3}{*}{3035} & RMSE & 98.09 &68.36  & 68.36 & \textbf{64.74} \\
& MAPE &15.15\% & 7.64\% &7.64\% & \textbf{7.59\%} \\
& MAE & 82.82 &51.15  &51.15  & \textbf{46.68}  \\
\hline 
\multirow{3}{*}{4004} & RMSE & 94.20  &61.89 &61.89 &\textbf{57.30}  \\
&MAPE & 16.44\%& 8.21\%& 8.21\%
& \textbf{6.80\%} \\
& MAE &80.25  &47.01  & 47.01 & \textbf{42.39} \\
\hline 
\multirow{3}{*}{4005} & RMSE & 106.9 & 87.65 &87.65 & \textbf{86.02} \\
&MAPE &10.02\% &6.34\% & 6.34\% & \textbf{6.27\%}
\\
& MAE &87.13 & 65.63 &65.63 & \textbf{64.35} \\
\hline 
\multirow{3}{*}{4050} & RMSE & 93.53 &57.85 &57.85 &\textbf{56.48} \\
&MAPE &37.96\% & 19.36\%& 19.36\%&\textbf{18.58\%}
\\
& MAE & 80.57 & 37.89 & 37.89 & \textbf{36.11} \\
\hline 
\multirow{3}{*}{4051} & RMSE &112.35  & 98.13 &98.13  &\textbf{98.60} \\
&MAPE &12.04\% &8.66\% & 8.66\% & \textbf{8.45\%} \\
& MAE & 83.94 & 65.56 & 65.56 & \textbf{65.20} \\
\hline 
\multirow{3}{*}{5062} & RMSE &90.63  & 60.06 & 60.06 &\textbf{56.18}  \\
&MAPE & 17.89\%& 8.23\%&8.23\% &\textbf{8.06\%} \\
& MAE &77.86  & 45.87 & 45.87 & \textbf{42.24} \\
\bottomrule

\end{longtable}

%\begin{figure*}
%\centering
%\subfigure[RMSE]{
% \includegraphics[width = 8cm]{RMSE.png}
% }
%\subfigure[MAPE]{
 %\includegraphics[width = 8cm]{MAPE.png}
% }\\
%\subfigure[MAE]{
 %\includegraphics[width = 10cm]{MAE.png}
% }
  %  \caption{CDF of RMSE, MAPE, MAE for different models}
   % \label{fig:evaluate}
%\end{figure*}
\par
Through Table \ref{tab:1}, we can find that DIDRN has more outstanding performance than other models. One layer LSTM has a much worse performance than other models since it is shallower. Deep LSTM seems to have a satisfying performance, however, on our machine, it actually takes us nearly 1 hour to train it before it converges. DRN shares a similar but lower error rate with deep LSTM and it is much easier to train. It only takes us about 20 minutes to train DRN. On the other hand, although DIDRN is derived from DRN, it has a lower error rate than DRN. In addition, it is not hard to train. We can finish training it within 30 minutes which is similar to DRN but is much shorter than that of deep LSTM. 
\par
\begin{figure*}
    \centering
    \includegraphics[width = 16cm]{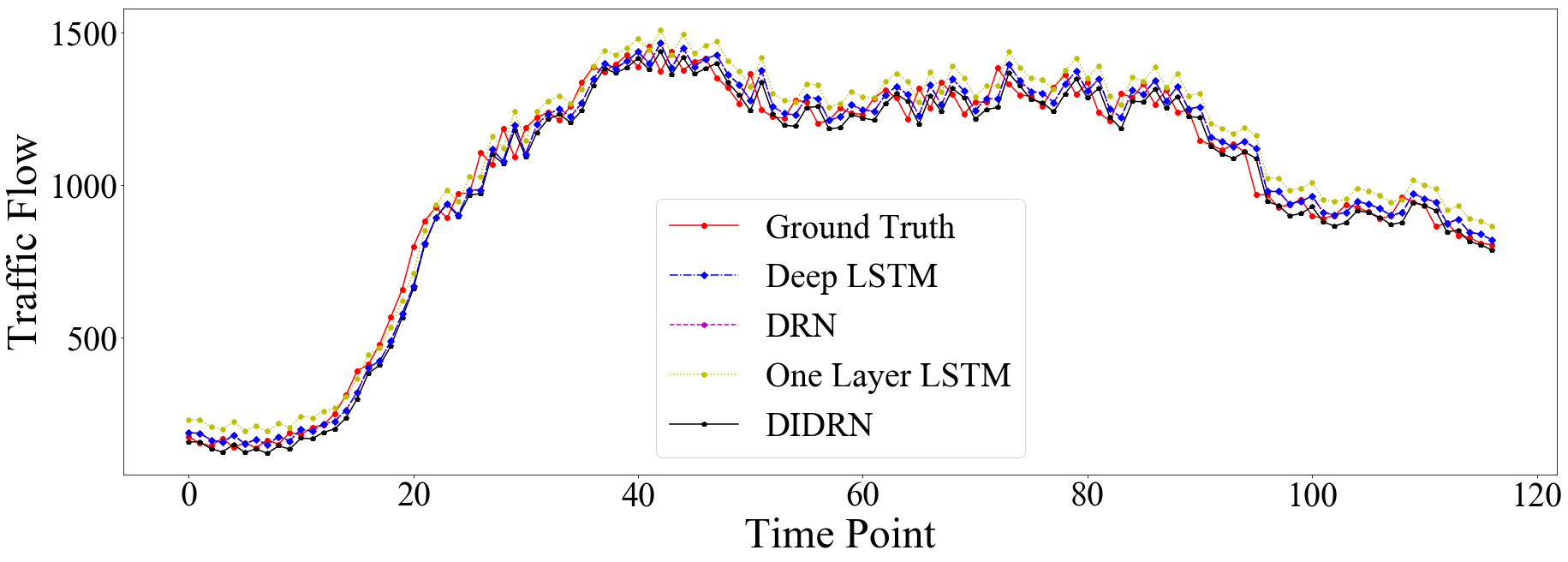}
    \caption{Predictions}
    \label{fig:com_pre}
\end{figure*}

Figure \ref{fig:com_pre} shows a comparison of predictions by different models.
\par
\begin{figure*}
    \centering
    \includegraphics[width = 16cm]{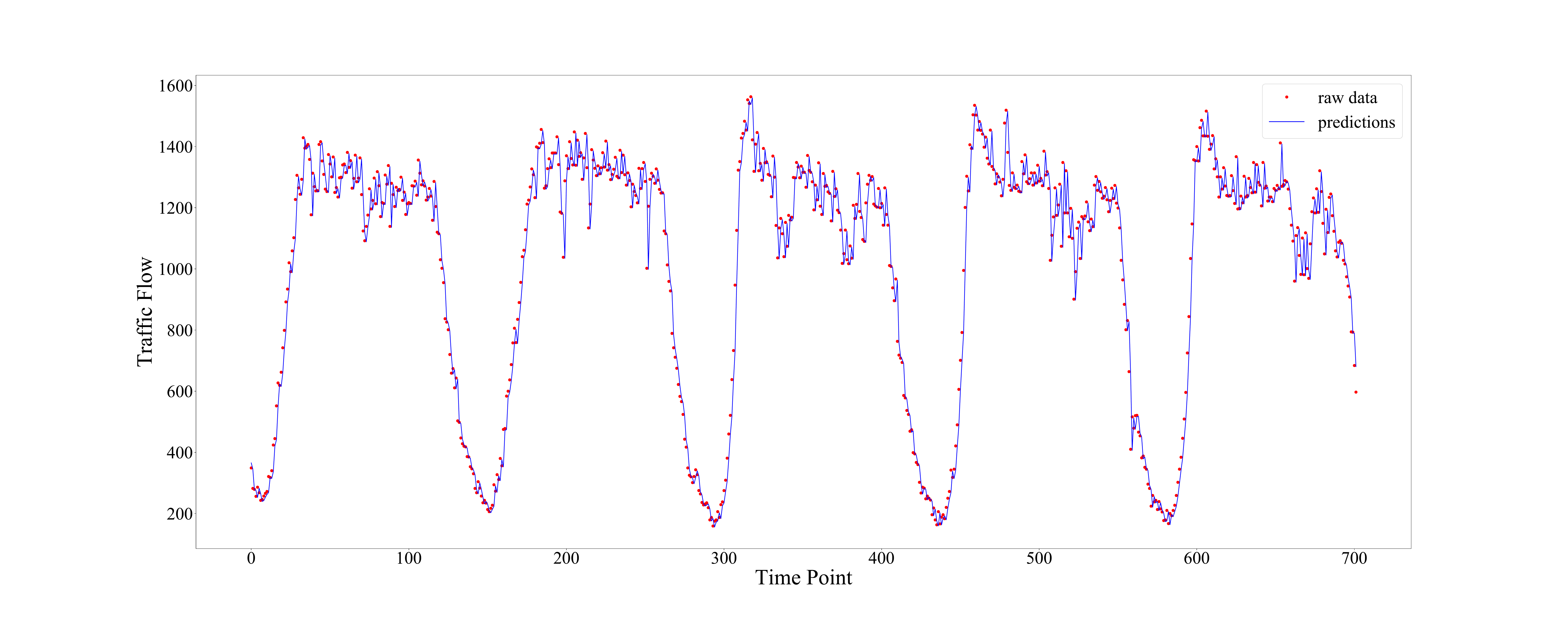}
    \caption{Raw data and predictions}
    \label{fig:pre}
\end{figure*}

Figure \ref{fig:pre} shows part of the predictions of DIDRN.

\par
\begin{figure*}
\centering
\subfigure[1 hour]{
 \includegraphics[width = 8cm]{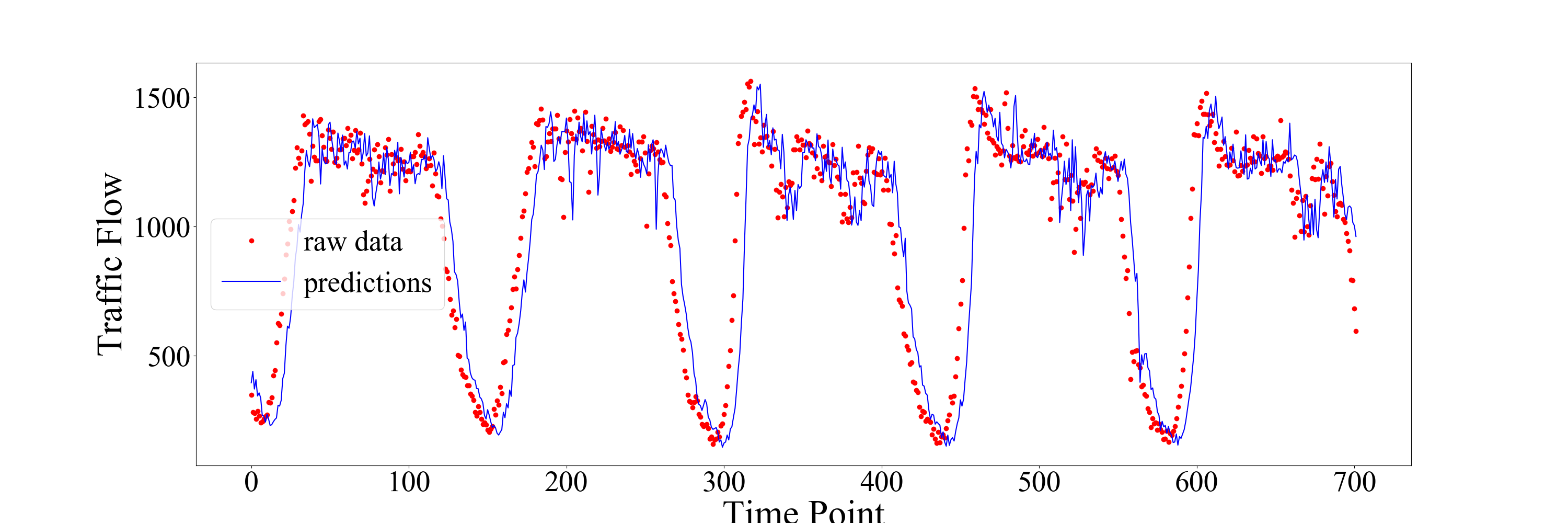}
 }
\subfigure[2 hours]{
 \includegraphics[width = 8cm]{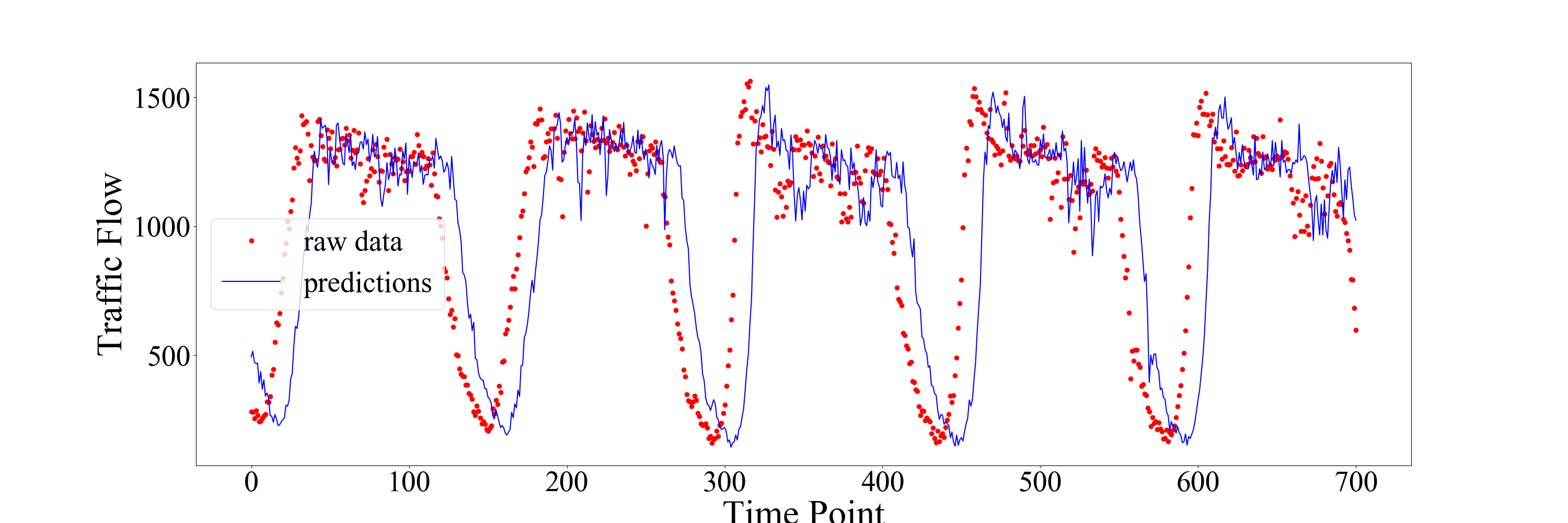}
 }\\
 \subfigure[24 hours]{
 \includegraphics[width = 10cm]{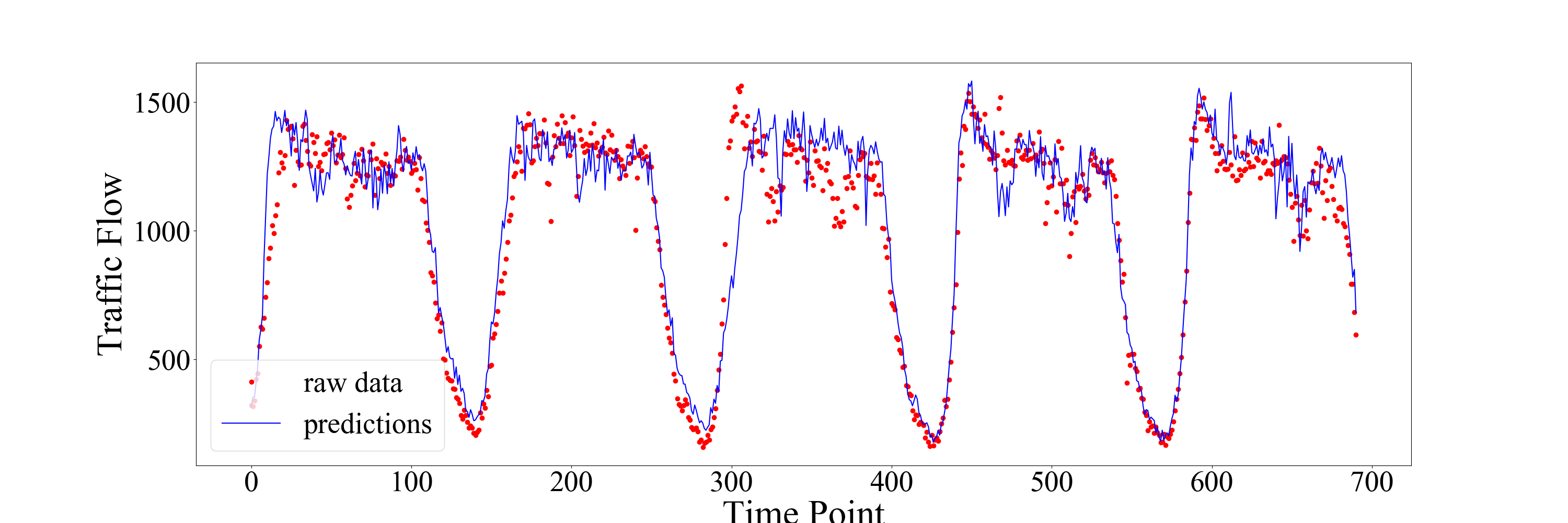}
 }
    \caption{Predictions of Different Time Intervals}
    \label{fig:pfdti}
\end{figure*}

To evaluate the stability of our proposed model, we change the time step into 1 hour, 2 hours, 24 hours and use DIDRN to conduct experiments.
Figure~\ref{fig:pfdti} and Table~\ref{tab:3} show our results.

\begin{table}
 \centering
    \caption{Performance of DIDRN with Different Time Intervals}
    \label{tab:3}
    \begin{tabular}{cccc}
    \toprule
    Time Interval (hour) & RMSE & MAPE & MAE\\
    \midrule
    1  & 202.87 & 18.45\% & 141.34 \\
    2  & 342.96 & 32.67\% & 233.05 \\
    24 & 135.24 & 10.88\% & 94.92 \\
    \bottomrule
    \end{tabular}
\end{table}
\par
In the following discussion, we conduct our analysis based on MAPE because it is simple and intuitive and the data of detector 2010 are chosen to demonstrate the results.
\par
From Table~\ref{tab:3} we can see that while the time interval becomes larger from 10 minutes to 1 hour and 2 hours, the accuracy decreases monotonously. This is congruent with our intuition. When the time interval becomes larger, we need more information to predict the traffic flow and the traffic flow of the previous time point is less related to the traffic flow of the predicted time point. Therefore, it is obvious that the accuracy would decrease. However, when we use the traffic flow of the previous day to predict that of the next day, the accuracy increase significantly to 10.88\% in terms of MAPE. This is not strange. On one hand, since our data are stable and have periodic nature, traffic flow of the same time in the previous day is quite similar to that of the predicted day. On the other hand, traffic flows 1 hour or 2 hours later can be really different with that of the current time point. For example, traffic flow at 7:00 a.m. has significant change compared to traffic flow at 8:00 a.m. and 9:00 a.m. since they are peak hours. People go to work during peak hours and consequently, lead to a sudden increase in traffic flow. The model do not have enough information to predict this sudden increase so the predict values have a larger deviation from the detected values.
\par
We explore the decrease and increase of accuracy when time interval increase from 10 minutes to 1 hour and from 1 hour to 24 hours in more detail.
\par
\begin{figure*}
    \centering
    \includegraphics[width = 14cm]{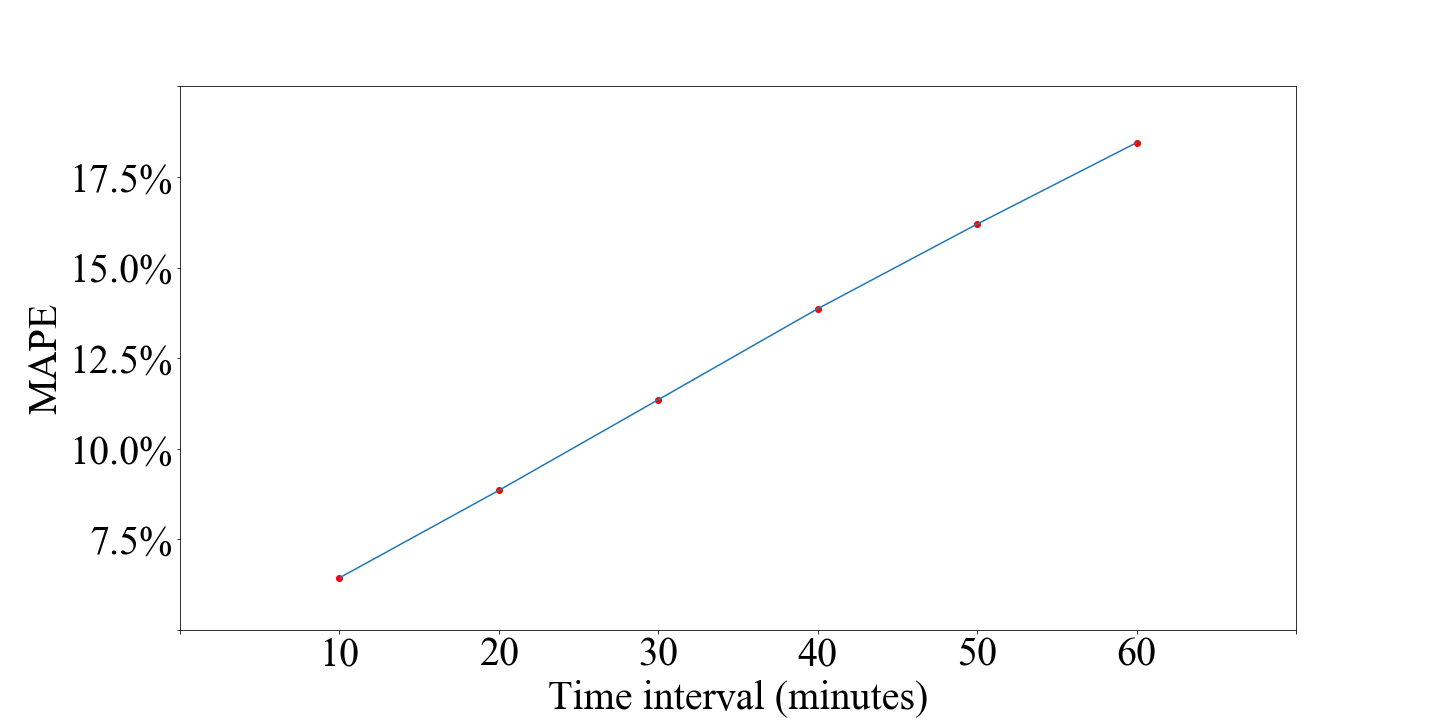}
    \caption{MAPE of Different Time Interval (minute)}
    \label{fig:minu}
\end{figure*}
\begin{figure*}
    \centering
    \includegraphics[width = 14cm]{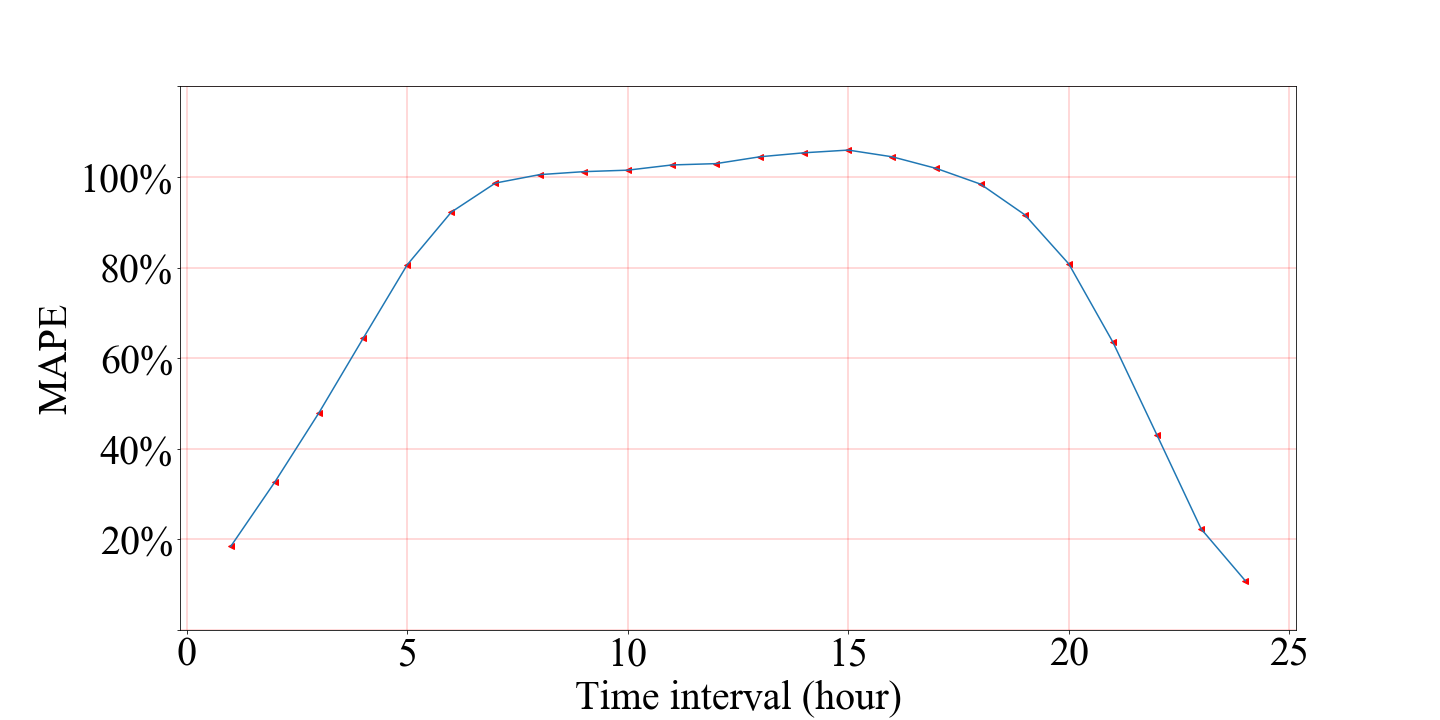}
    \caption{MAPE of Different Time Interval (hours)}
    \label{fig:hour}
\end{figure*}

Figure~\ref{fig:minu} shows the diagram of MAPE. It is obvious that MAPE is linearly related to time interval. This is true when the jump between time intervals is small, but when we increase the jump to 1 hour, this linear relation does not exist any more. Figure~\ref{fig:hour} shows the MAPE of DIDRN when time interval increase from 1 hour to 24 hours. We can observe that the shape of this curve is like a pudding. MAPE firstly increase from about 18\% to over 100\% then decrease to less than 11\%. When time interval equals to 2 hours, the error rate is unacceptable. The highest error rate is over 105\% which means the models almost cannot estimate the traffic flow when time interval is large. The reason is that we only use the current traffic flow to estimate traffic flow of the next time point. Traffic flow would not have significant change when time interval is not too large. However, when time interval is as large as several hours, the current traffic flow is much different with the traffic flow to predict. This kind of time interval is equivalent to a reshuffle. The specific example can be what we have mentioned above.
\par
In order to further analyze the performance of our model, we gradually increase the time interval to 7 days. 
\par
Table~\ref{tab:4} shows the result. It can be observed that when the time interval is ranging from 2 days to 6 days, the error rate decreases from 13.96\% to 10.45\%. This is because when time interval becomes larger, the predicted traffic flow has less relation with the input traffic flow. The model has less information to update its weights in order to perform well both in training set and test set.
\par
There is a big reduction between 1 day and 2 days. If we think carefully about the whole prediction process, we can figure out what has happened. When time interval equals to 1 day, we can find that traffic flow of Friday and Saturday are used to predict traffic flow of Saturday and Sunday, respectively. Nevertheless, when time interval equals to 1 days, traffic flow of Friday and Saturday are used to predict traffic flow of Sunday and Monday, respectively. Saturday and Sunday are weekend and Monday and Friday are workdays. As less people will go to work at weekends, traffic flow of weekend and workday are significantly different. Hence, it is not hard to see that when time interval equals to 2 days, the accuracy will decrease to some extent.
\par
There is a large increase between 6 days and 7 days. Due to symmetry, the error rate of 6 days should be similar with the error rate of 1 day and it can be observed from our result too. Since traffic flow are a kind of periodic data and the period is exact 7 days, it is not surprising that the flow one week later is similar with the current flow. Due to aforementioned property, our model can perform better and thus has a higher accuracy.

\begin{table}
 \centering
    \caption{Performance of DIDRN with Different Days}
    \label{tab:4}
    \begin{tabular}{cccc}
    \toprule
    Days & RMSE & MAPE & MAE\\
    \midrule
    1 &135.24 &10.88\% & 94.92 \\
    2 &158.27 &13.96\% &113.70 \\
    3 &154.55 &13.81\% &109.36 \\
    4 &145.62 &13.22\% &104.54 \\
    5 &142.73 &10.85\% &99.14 \\
    6 &143.36 &10.45\% &98.12 \\
    7 &100.56 &7.28\% &69.48 \\
    \bottomrule
    \end{tabular}
\end{table}
\begin{figure*}

\centering
\subfigure[Detector 2010]{
\includegraphics[width = 8cm]{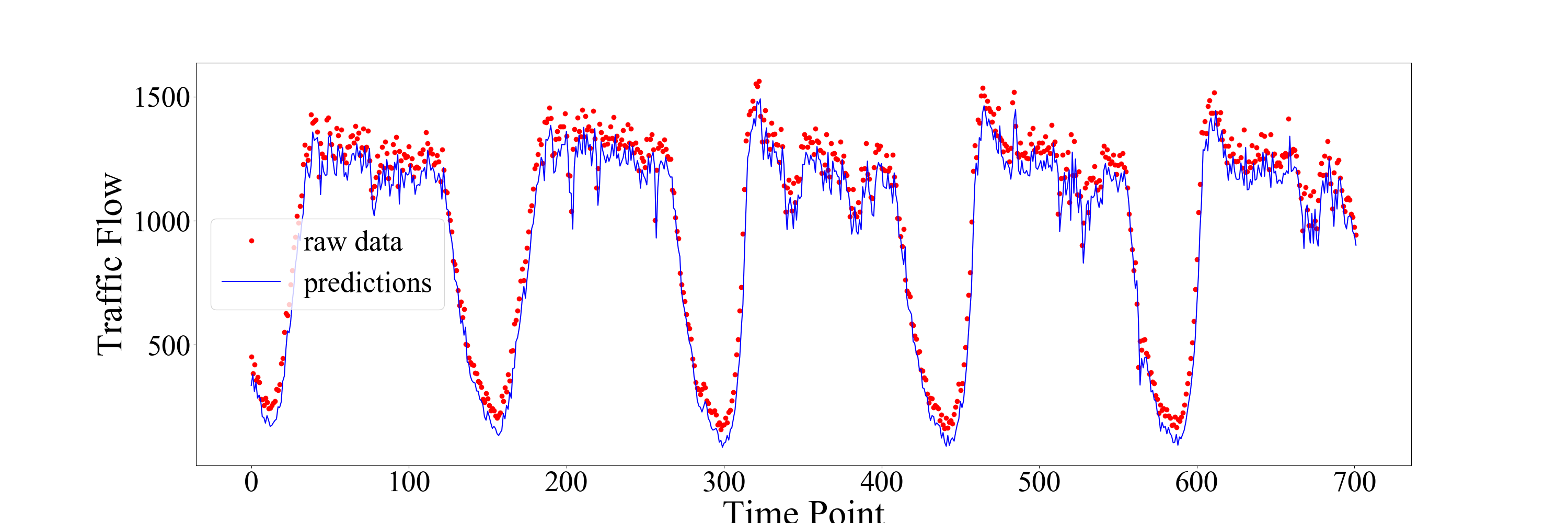}
}
\subfigure[Detector 2011]{
\includegraphics[width = 8cm]{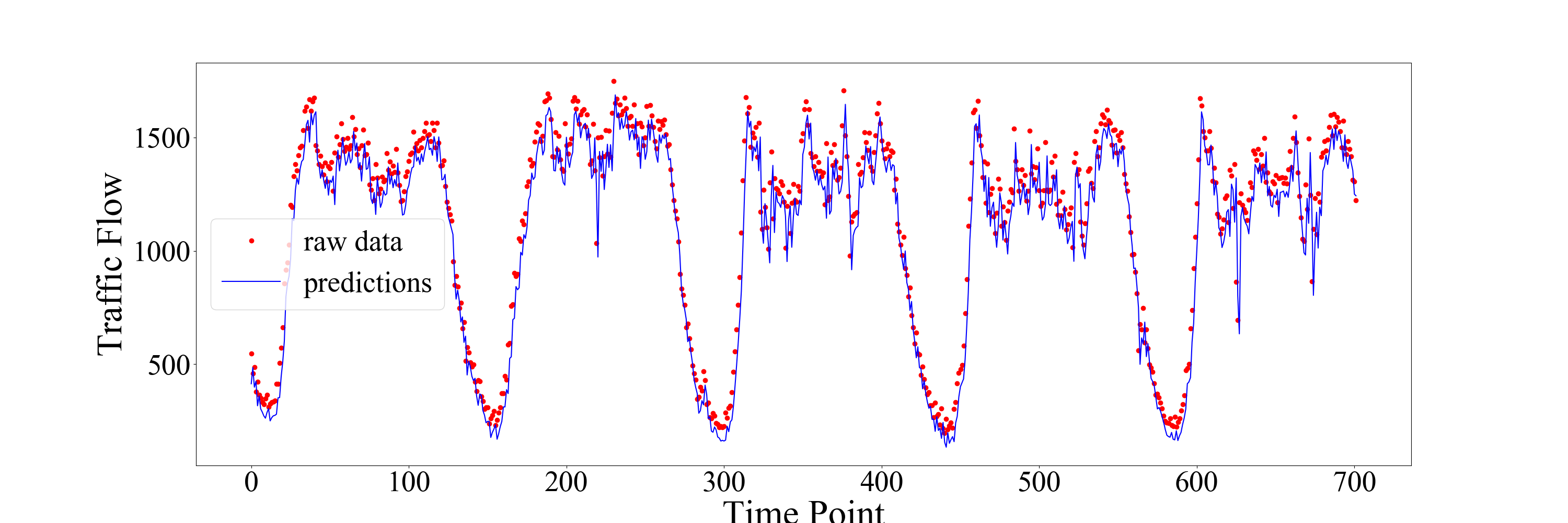}
}\\
\subfigure[Detector 2012]{
\includegraphics[width = 8cm]{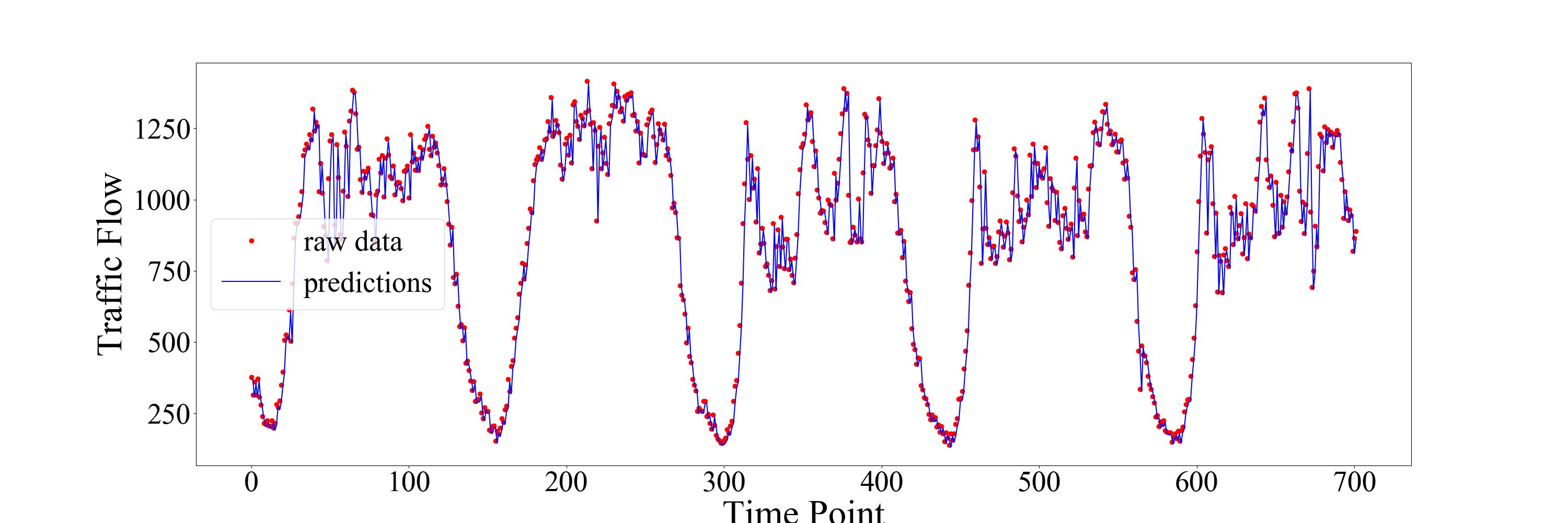}
}
\subfigure[Detector 2023]{
\includegraphics[width = 8cm]{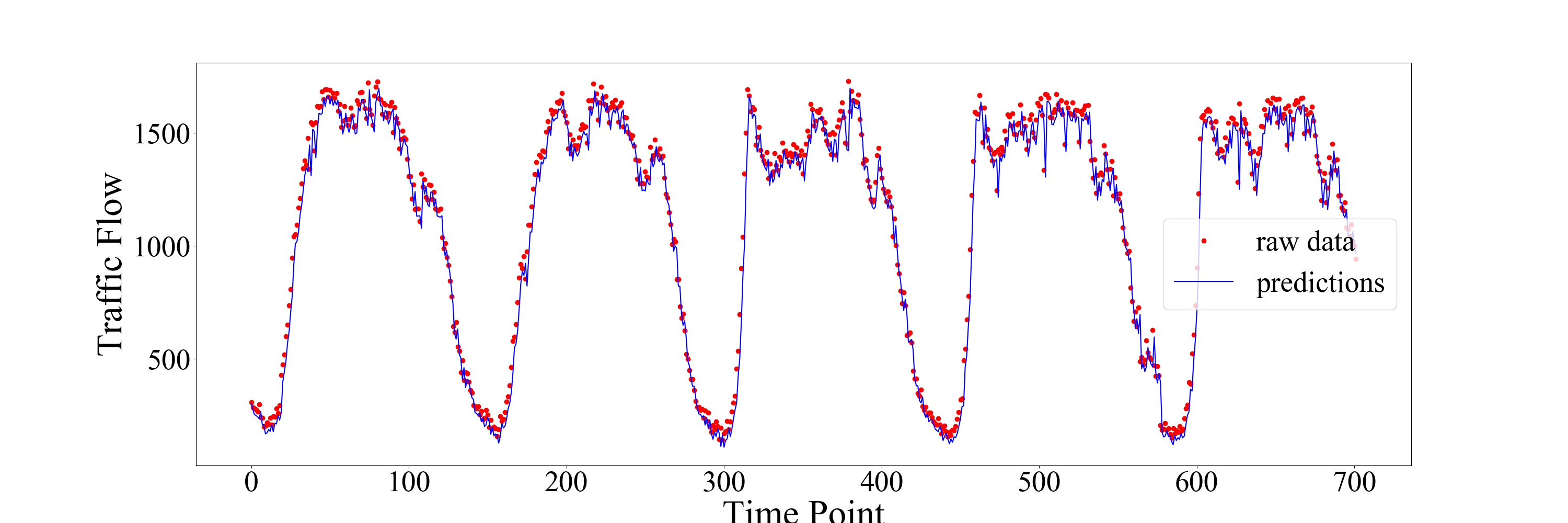}
}\\
\subfigure[Detector 2030]{
\includegraphics[width = 8cm]{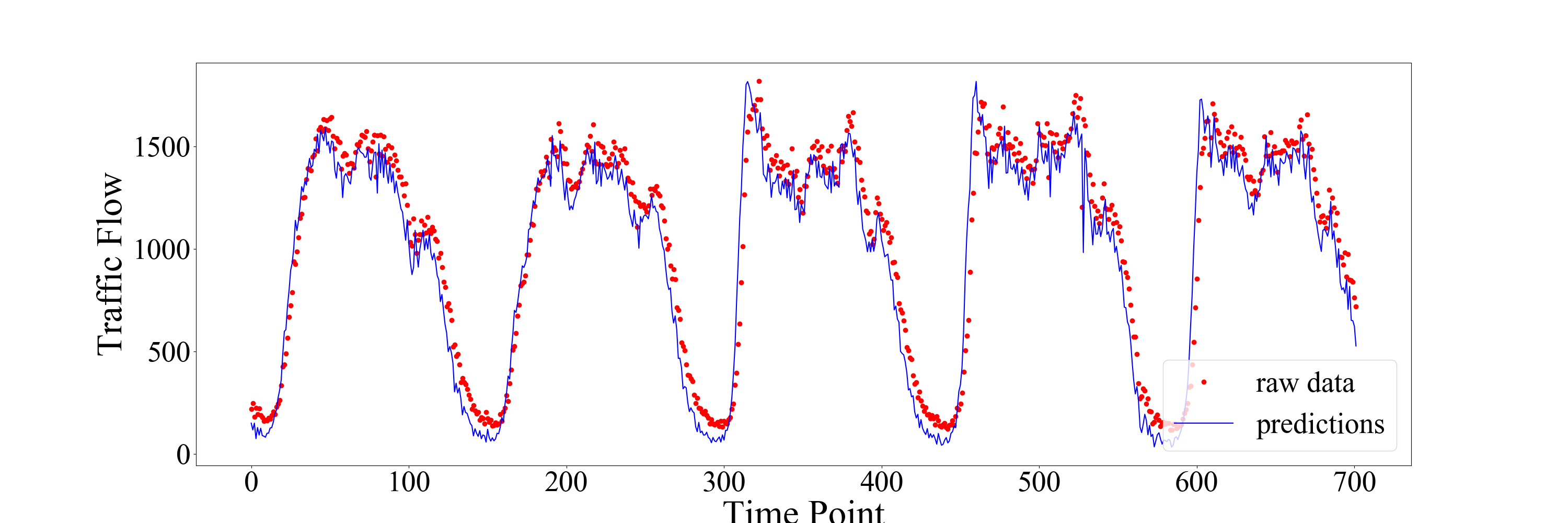}
}
\subfigure[Detector 2033]{
\includegraphics[width = 8cm]{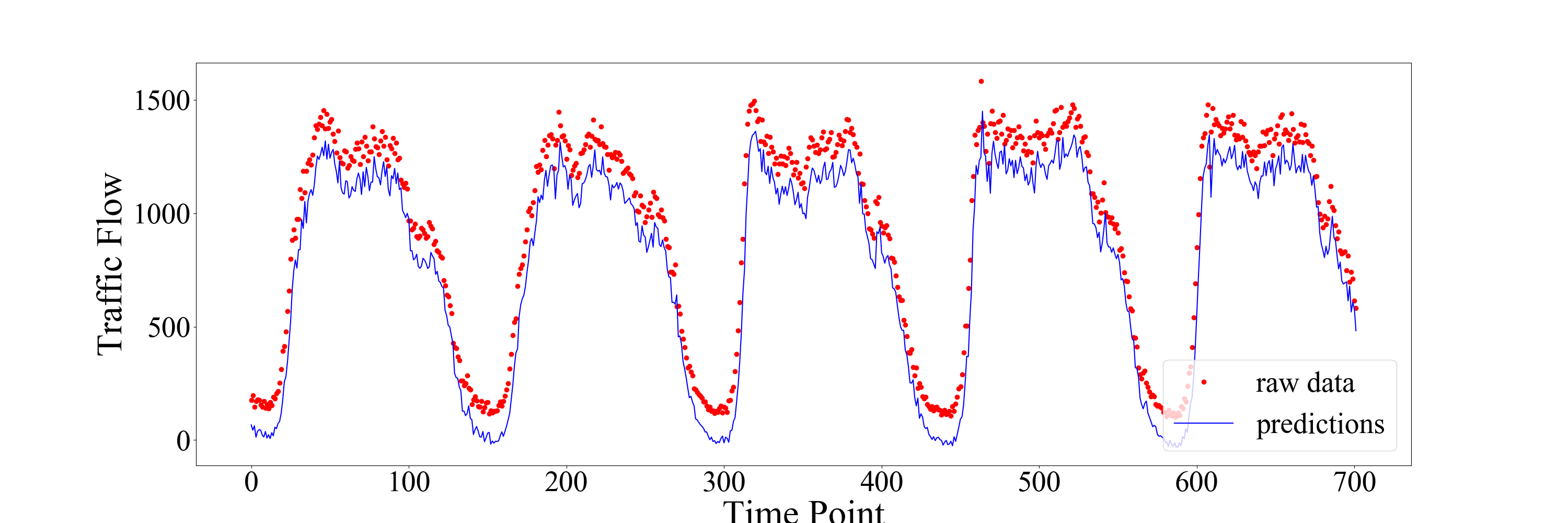}
}\\
\subfigure[Detector 2052]{
\includegraphics[width = 8cm]{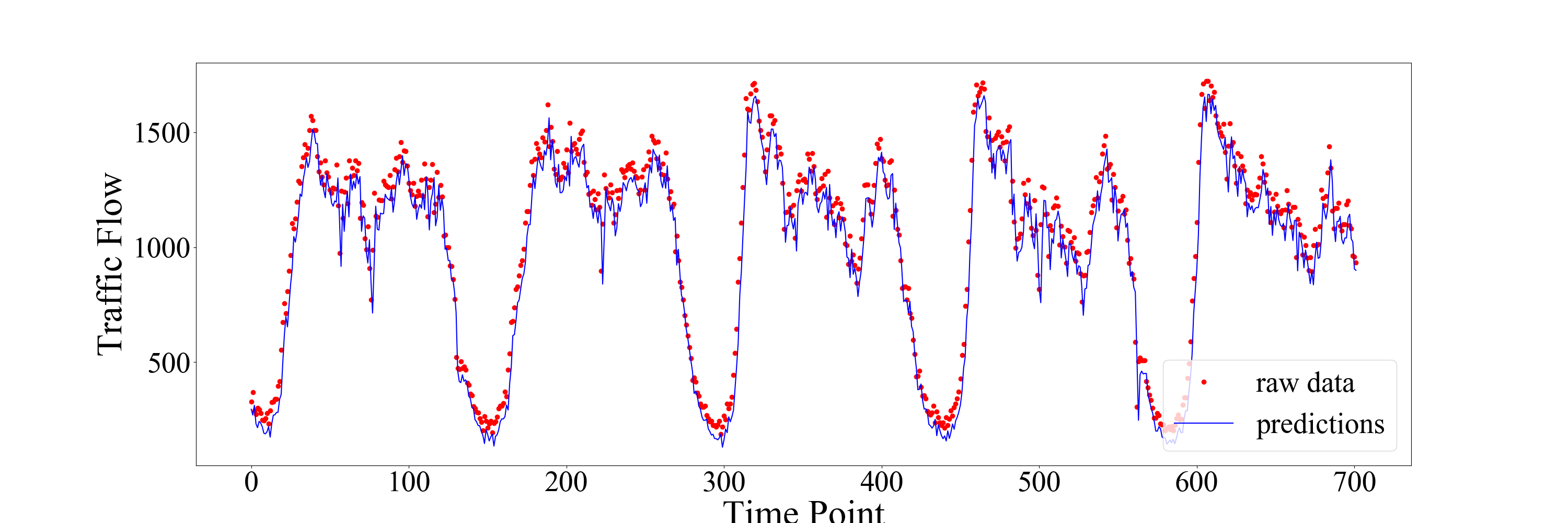}
}
\subfigure[Detector 3034]{
\includegraphics[width = 8cm]{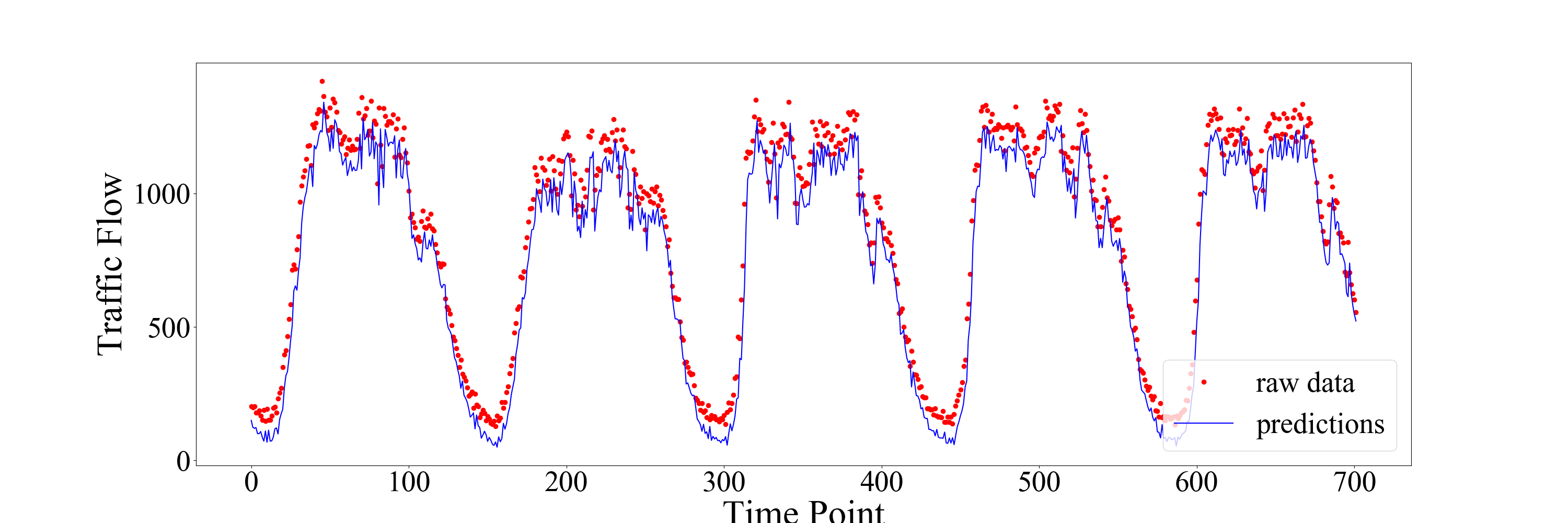}
}\\
\subfigure[Detector 3035]{
\includegraphics[width = 8cm]{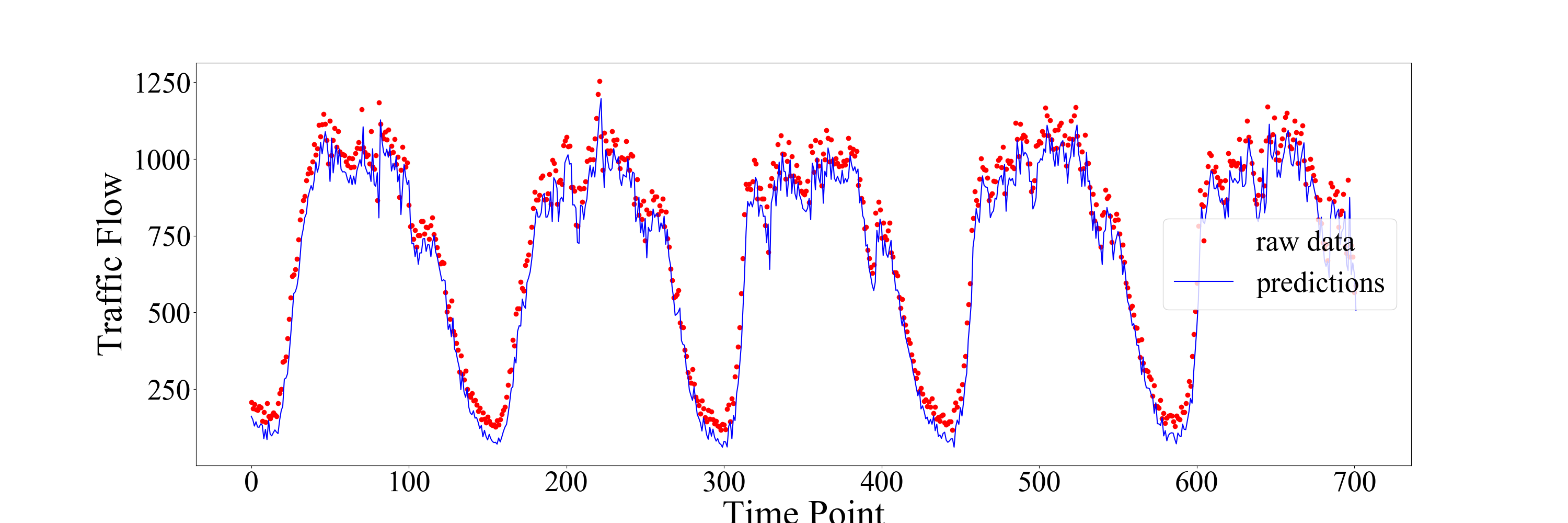}
}
\subfigure[Detector 4004]{
\includegraphics[width = 8cm]{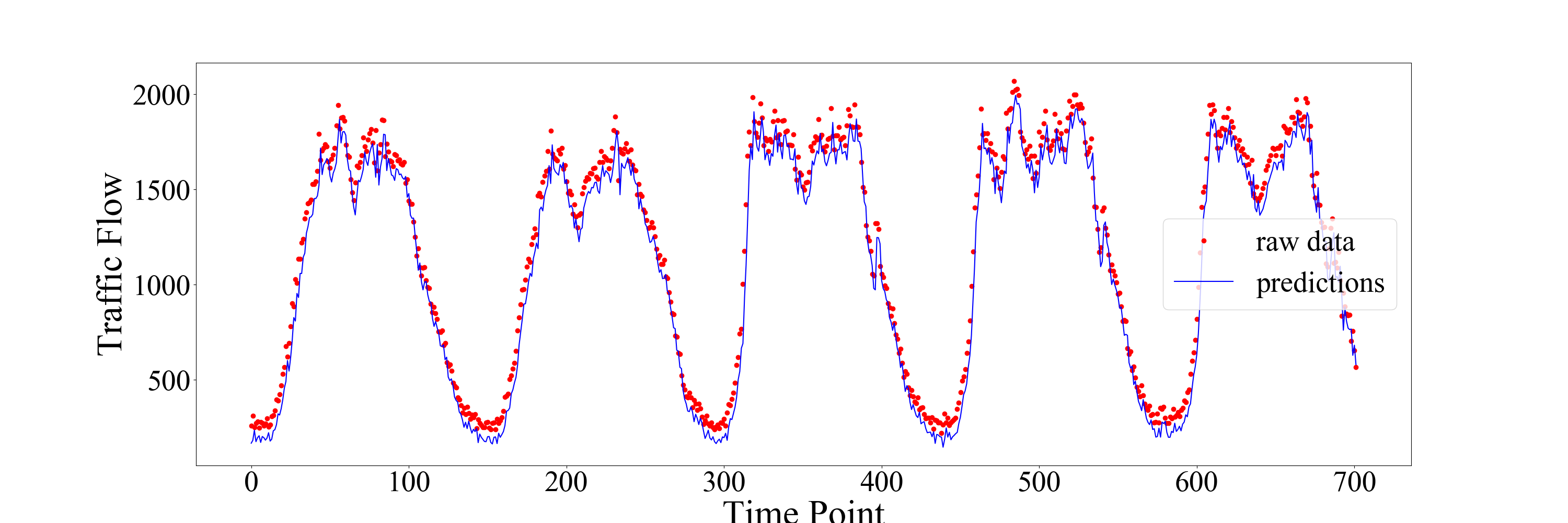}
}\\
\subfigure[Detector 4005]{
\includegraphics[width = 8cm]{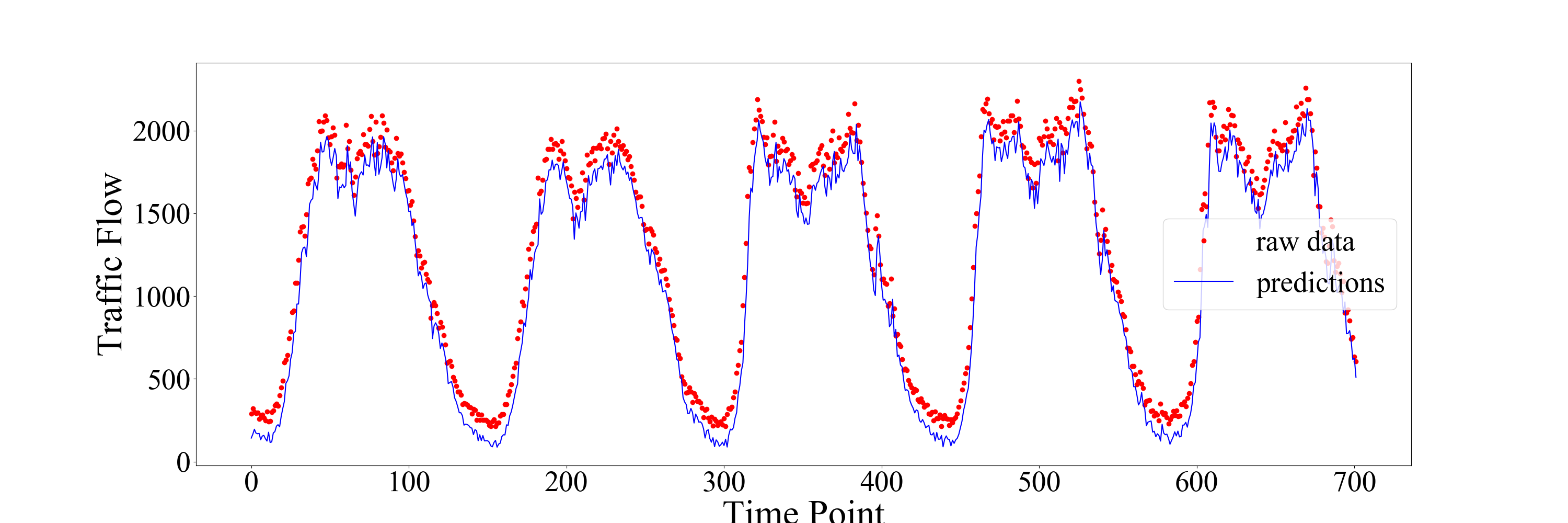}
}
\subfigure[Detector 4050]{
\includegraphics[width = 8cm]{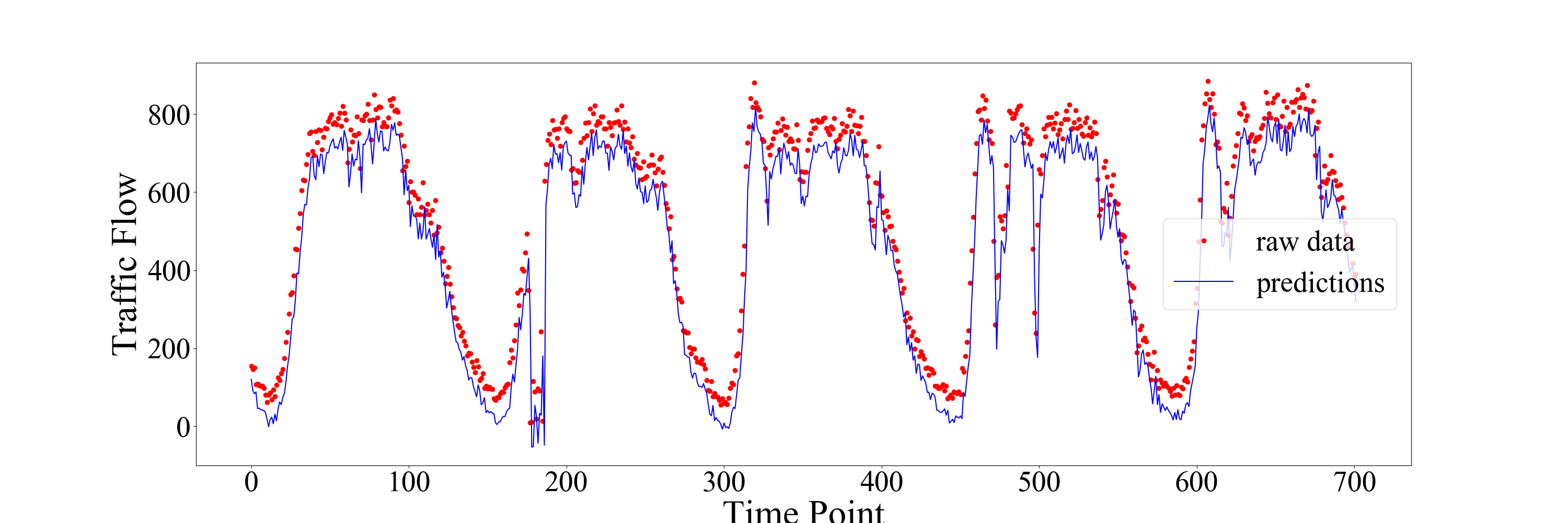}
}\\
\subfigure[Detector 4051]{
\includegraphics[width = 8cm]{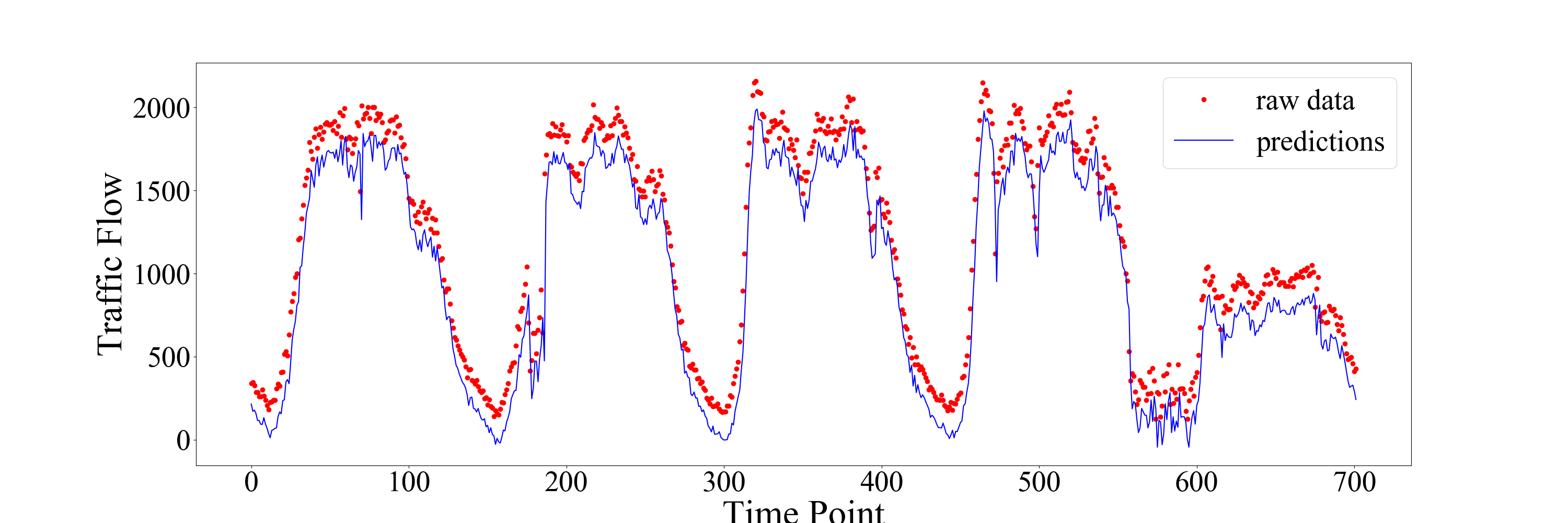}
}
\subfigure[Detector 5062]{
\includegraphics[width = 8cm]{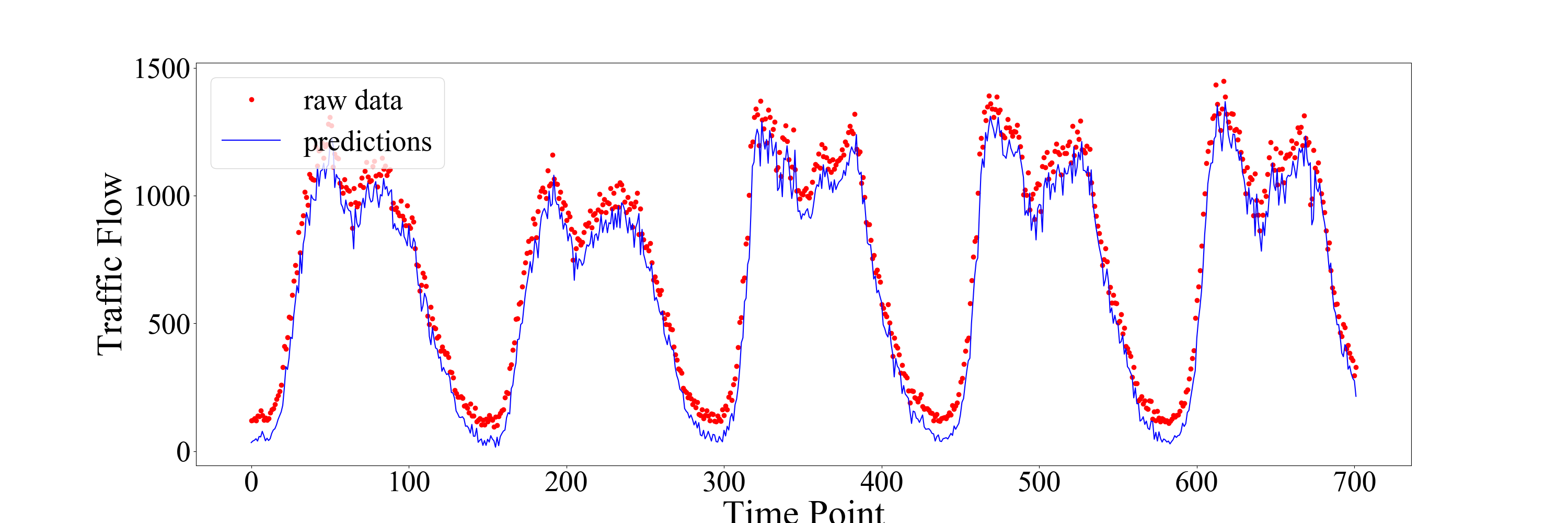}
}\\
\end{figure*}
In summary, our model can do well in short-term traffic flow prediction as we use traffic flow at one time point to predict traffic flow at next time point. If we use more information as input, that is, use not only the current traffic flow but also that of previous one or more time points, our model is likely to perform better.
\section{Conclusion and Future work} \label{sec:confw}

In this paper, we demonstrate the basic ideas of deep residual network. It turns out that DRN is much simpler to train and have an excellent performance. Then we explain how we are inspired by DRN and and how we improve it to do our research. We show the architecture of our network and how it works. After that, we propose our dynamic model DIDRN and demonstrate why it makes sense.
\par
We show the entire process of processing data step by step. We explain how we process the raw data into data that can be used in supervised learning and how we get the final predictions. Next, we develop several experiments and compare performance of different models. It turns out that DIDRN has better performance than some popular models. From the view of MAPE, DIDRN has a 1.41\% performance improvement at most comparing to LSTM and DRN.  For RMSE and MAE, DIDRN can  have a mostly 9 of reduction.
\par
Despite the good performance, our model still have some shortcomings.To summarize, our main contributions are as follows:
\begin{itemize}
    \item We apply deep residual network in traffic flow prediction and improve it.
    \item We take practical applications into account and propose a dynamic model called DIDRN.
    \item The results show that our model is more powerful than other commonly used models.
\end{itemize}

\par
We only take temporal pattern into account in this paper. In future work, the spatial pattern would be considered and we will make our model learn spatial-temporal dependence. 
\par
In addition, our model can do traffic flow prediction well when the time interval is short or is the period of our data. But it has a poor performance when the time interval is a little larger. Future work can be done to extend the model to a more generalized version so that the model can perform well when time interval is both short and large.
\par
Moreover, since weather condition certainly has an impact on traffic flow, it would be considered that making our network learn the correlation between weather condition and traffic flow.
\bibliographystyle{plainnat}
\bibliography{Reference.bib}
\end{document}